\documentclass[twocolumn,times,final]{elsarticle}

\usepackage{amssymb}
\usepackage{latexsym}
\usepackage{url}
\usepackage{amsmath}
\usepackage{hyperref}
\hypersetup{
    colorlinks=true,
    linkcolor=blue,
    filecolor=blue,
    urlcolor=blue,
    citecolor=blue,
}
\usepackage{tabularx}
\usepackage{setspace}
\usepackage{soul}
\usepackage{booktabs}
\usepackage{adjustbox}
\usepackage{changepage}
\usepackage{ragged2e}

\journal{arXiv}

\begin{document}

\begin{frontmatter}

\title{Adaptive Law-Based Transformation (ALT): A Lightweight Feature Representation for Time Series Classification}

\author[1,2]{Marcell T. Kurbucz\corref{cor1}}
\cortext[cor1]{Corresponding author:
Tel.: +36-1-392-2222}
\ead{kurbucz.marcell@wigner.hun-ren.hu}

\author[1,3]{Bal\'azs Haj\'os}
\author[1,4]{Bal\'azs P. Halmos}
\author[1,4]{Vince \'A. Moln\'ar}
\author[1,2]{Antal Jakov\'ac}

\address[1]{Department of Computational Sciences, Wigner Research Centre for Physics, 29-33 Konkoly-Thege Miklós Street, Budapest, 1121, Hungary}
\address[2]{Department of Statistics, Corvinus University of Budapest, 8 Fővám Square, Budapest, 1093, Hungary}
\address[3]{Faculty of Science, Eötvös Loránd University, 1/A Pázmány Péter Walkway, Budapest, 1117, Hungary}
\address[4]{Faculty of Engineering and Natural Sciences, Tampere University, Kalevantie 4, Tampere, 33100, Finland}

\begin{abstract}
Time series classification (TSC) is fundamental in numerous domains, including finance, healthcare, and environmental monitoring. However, traditional TSC methods often struggle with the inherent complexity and variability of time series data. Building on our previous work with the linear law-based transformation (LLT)---which improved classification accuracy by transforming the feature space based on key data patterns---we introduce adaptive law-based transformation (ALT). ALT enhances LLT by incorporating variable-length shifted time windows, enabling it to capture distinguishing patterns of various lengths and thereby handle complex time series more effectively. By mapping features into a linearly separable space, ALT provides a fast, robust, and transparent solution that achieves state-of-the-art performance with only a few hyperparameters.
\end{abstract}


\end{frontmatter}

\section{Introduction}
\label{sec:introduction}

\noindent
Time series classification (TSC) is essential in various domains such as finance, healthcare, and environmental monitoring, where the goal is to categorize temporal data into predefined classes~\cite{esling2012time,fawaz2019deep}. Traditional TSC approaches often rely on feature extraction methods designed to capture the temporal dynamics and structural patterns inherent in time series data~\cite{agrawal1993efficient,box2015time,fulcher2014highly}. However, these methods may struggle with the complexities and variability of time series data.

Our previous work introduced the linear law-based transformation (LLT) method, which performs uni- and multivariate TSC tasks by transforming the feature space based on identified governing patterns in the data~\cite{kurbucz2022facilitating}. LLT uses time-delay embedding and spectral decomposition to extract linear laws from training data and applies these laws to transform test data, resulting in improved classification accuracy with low computational cost.

In this paper, we build upon the LLT method by introducing an enhanced approach called adaptive law-based transformation (ALT) that utilizes variable-length shifted time windows. Unlike LLT, which operates on fixed-length windows, ALT explores patterns of varying lengths and shifts, making it more effective in capturing distinguishable patterns within time series data. This flexibility allows the method to identify local patterns of different scales, enhancing its ability to classify complex time series.

Similar to LLT, our method aims to transform features into a linearly separable feature space, offering a fast, robust, and transparent solution that achieves state-of-the-art performance. By reducing the need for extensive hyperparameter tuning and incorporating variable-length patterns, ALT simplifies the modeling process and enhances interpretability, setting it apart from mainstream neural networks and other deep learning techniques.

We evaluated ALT on eleven benchmark time series datasets, demonstrating its effectiveness compared to existing TSC techniques, including the original LLT method. The results show that the proposed approach not only achieves higher accuracy but also offers advantages in speed and transparency.

The remainder of this paper is organized as follows: Section~\ref{sec:related_work} reviews related work, including the LLT method. Section~\ref{sec:method} describes the datasets used and details our proposed method. Section~\ref{sec:results} presents and discusses the experimental results. Finally, Section~\ref{sec:conclusions} concludes the paper and suggests directions for future research.

\noindent
\section{Related Work}
\label{sec:related_work}

\noindent
Time series classification (TSC) methods can generally be grouped into three main categories: feature-based, distance-based, and deep learning-based approaches. Each category offers distinct advantages and faces specific challenges.

Feature-based approaches extract meaningful representations from time series data before applying classification algorithms. These representations may capture statistical descriptors~\cite{fulcher2014highly}, spectral transformations such as the discrete Fourier transform (DFT) or discrete wavelet transform (DWT)~\cite{agrawal1993efficient}, or model-based features derived from techniques like autoregressive integrated moving average (ARIMA)~\cite{box2015time}. Shapelet-based methods, which identify short, discriminative subsequences (shapelets) within the data~\cite{ye2009time}, can be considered a subset of feature-based methods~\cite{ji2019just}. Shapelet-based approaches focus on local features that are highly interpretable and often effective for capturing localized variations, though they may struggle with multi-scale patterns and can be computationally intensive for long time series. Feature-based representations are typically classified using conventional methods such as logistic regression, random forests, or support vector machines (SVM).

Distance-based methods measure the similarity or dissimilarity between entire time series without explicitly transforming them into feature vectors. A well-known example is dynamic time warping (DTW)~\cite{senin2008dynamic}, which is robust to local temporal distortions and useful for aligning time series. However, these methods can become computationally expensive as the dataset size grows, and they lack an interpretable intermediate representation of the data.

Deep learning-based methods automatically learn hierarchical feature representations directly from raw time series. Convolutional neural networks (CNNs) are adept at identifying local temporal correlations, while recurrent neural networks (RNNs) excel at capturing sequential patterns, including long-term dependencies~\cite{fawaz2019deep, karim2019multivariate, zheng2014time}. While deep learning methods often achieve strong empirical performance, they typically require large labeled datasets, involve extensive hyperparameter tuning, and may lack transparency in their learned representations.

The linear law-based transformation (LLT)~\cite{kurbucz2022facilitating} integrates elements of feature-based and distance-based methods. By using time-delay embedding and spectral decomposition, LLT extracts governing patterns from training data and applies these patterns to unseen instances, transforming the feature space to improve classification accuracy. Despite its low computational cost, LLT relies on fixed-length windows, which can limit its ability to capture patterns of variable lengths.

Building on LLT, this work introduces the adaptive law-based transformation (ALT). ALT incorporates variable-length shifted time windows to capture local patterns across multiple temporal scales while maintaining interpretability and computational efficiency. This adaptive design enables ALT to effectively handle complex time series, bridging the gaps between diverse TSC approaches.

\section{Data and Methodology}
\label{sec:method}

\subsection{Employed Data}
\label{sec:2.1}

\noindent
This study utilizes eleven real-world datasets sourced from the UCR Time Series Classification Archive \cite{dau2019ucr, UCRArchive2018}.\footnote{These datasets are available at: \url{https://www.timeseriesclassification.com} (retrieved: January 15, 2025).} The datasets are detailed in Table~\ref{tab:datasets}.

\begin{table*}[ht]
    \centering
    \caption{Overview of the datasets employed in this study}
    \resizebox{\textwidth}{!}{
    \label{tab:datasets}
    \begin{tabular}{lcccccccp{13cm}}
        \toprule
        \textbf{Dataset} & \textbf{Type} & \textbf{Classes} & \textbf{Features} & \textbf{Train Size} & \textbf{Test Size} & \textbf{Length} & \textbf{Balanced} & \multicolumn{1}{c}{\textbf{Description}} \\ \midrule
        BasicMotions & Multivariate & 4 & 6 & 40 & 40 & 100 & Yes & Contains motion sensor data from four different activities performed by participants. \\ 
        Coffee & Univariate & 2 & 1 & 28 & 28 & 286 & Yes & Spectrographs of two types of coffee beans, with the task of differentiating between them. \\
        Epilepsy & Multivariate & 4 & 3 & 137 & 138 & 207 & Yes & Data collected from a tri-axial accelerometer while participants performed four tasks, including mimicking a seizure. \\
        Epilepsy2 & Univariate & 2 & 1 & 80 & 11420 & 178 & \vtop{\hbox{\strut Train}\hbox{\strut ~only}} & Single-channel EEG measurements aimed at determining whether a participant is experiencing a seizure. \\
        FordA & Univariate & 2 & 1 & 1320 & 3601 & 500 & Yes & Measurements of engine noise in automotive production, used to detect specific symptoms. \\
        FordB & Univariate & 2 & 1 & 810 & 3636 & 500 & Yes & Similar to FordA, but focuses on detecting different symptoms in engine noise measurements. \\
        GunPoint1 & Univariate & 2 & 1 & 50 & 150 & 150 & Yes & This dataset records X-axis hand motions for ``Gun-Draw'' and ``Point'' actions by two actors. \\
        GunPoint2 & Univariate & 2 & 1 & 135 & 316 & 150 & Yes & Variation of the GunPoint dataset focusing on distinguishing participants from different age groups. \\
        GunPoint3 & Univariate & 2 & 1 & 135 & 316 & 150 & Yes & Variation of the GunPoint dataset focusing on distinguishing male and female participants. \\
        GunPoint4 & Univariate & 2 & 1 & 135 & 316 & 150 & Yes & Variation of the GunPoint dataset focusing on distinguishing old and young participants. \\
        PowerCons & Univariate & 2 & 1 & 180 & 180 & 144 & Yes & Device power consumption data, with the task of determining the operational status. \\
        \bottomrule
        \multicolumn{9}{l}{\textbf{Note:} The original names of the GunPoint datasets, marked by numbers, are as follows: 1. GunPoint; 2. GunPointAgeSpan; 3. GunPointMaleVersusFemale; 4. GunPointOldVersusYoung.}
    \end{tabular}
    }
\end{table*}

\subsection{Feature Representation and Classification}
\label{ssec:formalization}

\noindent A general TSC task can be formalized as follows. The input data is represented as $x_t^{i,j}$, where $t \in {1, 2, \dots, h}$ denotes the observation times, $i \in {1, 2, \dots, \tau}$ identifies the instances, and $j \in {1, 2, \dots, m}$ indexes the different input series belonging to a given instance. The output $y^i \in {1, 2, \dots, c}$ identifies the class of instance $i$. The task is to predict the classes from the input data. To address this task, we use the following algorithm:

\begin{adjustwidth}{2pt}{0pt}

\begin{enumerate}[noitemsep]

\item[\textbf{[A1]}] \textbf{Data Splitting.} Divide the instances into learning ($Lr$), training ($Tr$), and test ($Te$) subsets using random selection stratified by class representation.

\item[\textbf{[A2]}] \textbf{Sequence Extraction.} (For each $Lr$, $j$, and $(r, l, k)$): Extract $r$-length sequences using shifted time windows (shifted by $k$), and take out $2l-1$ points evenly. The triplets $(r, l, k)$ are pre-defined parameters, where $r\leq h$, and $(2l-2) \mid (r-1)$. For a given $Lr$, $j$, and $(r, l, k)$, $ \left \lfloor{\frac{h - r + 1}{k}}\right \rfloor$ sequences are generated.

\item[\textbf{[A3]}] \textbf{Shapelet Vectors.} (For each sequence): Perform $l$-dimensional time-delay embedding \citep{takens1981dynamical} ($S$)—where $2l-1$ denotes the length of the given sequence, and $S$ is a symmetric matrix. Perform spectral decomposition of $S$. The eigenvector for the smallest absolute eigenvalue ($\in \mathbb{R}^+ \cup \{0\}$) is called the $v$ shapelet vector, and $Sv \approx 0$.\footnote{Note that this step relates to principal component analysis (PCA) \citep{gao2021human}, which extracts informative directions using eigenvectors of the largest eigenvalues. In contrast, ALT focuses on the dimension where $S$ shows the least variability, using the corresponding $v$ vector to compare shapelets.}\\

\item[\textbf{[A4]}] \textbf{Shapelet Matrices.} (For each $j$ and $(r, l, k)$): Use shapelet vectors related to the same $j$ and $(r, l, k)$ pairs as the column vectors of the shapelet matrix $P$. Group patterns based on the related class within $P$ ($c$ classes result in $c$ partitions within the $P$ matrix).

\item[\textbf{[A5]}] \textbf{Transformation.} (For each $Tr$, $j$, and $(r, l, k)$): Let $s = \frac{r - 1}{2l - 2}$ and $o = \left\lfloor \frac{h - sl + 1}{k} \right\rfloor$. Embed the instance into an $o \times l$ matrix $(A)$ as follows:
\hspace*{\fill}
\begin{equation}
    A = 
    \begin{pmatrix}
    x_1^{i, j} & x_{s+1}^{i, j} & \dots  & x_{(l-1)s+1}^{i, j} \\
    x_{k+1}^{i, j} & x_{k+s+1}^{i, j} & \dots & x_{k+(l-1)s+1}^{i, j} \\
    \vdots & \vdots & \ddots & \vdots \\
    x_{(o-1)k+1}^{i, j} & x_{(o-1)k+s+1}^{i, j} & \dots & x_{(o-1)k+(l-1)s+1}^{i, j} \\
    \end{pmatrix}.
\end{equation}
Right-multiply this matrix with the $P$ shapelet matrix related to the same pair of $j$ and $(r, l, k)$, that is, $O = AP$. Shapelets from each class in $P$ ``compete'' to transform the $A$ matrix close to null vectors.

\item[\textbf{[A6]}] \textbf{Feature Generation.} (For each transformed matrix): Square the values of the resulting $O$ and partition it by the class from which the shapelets originate. Different methods are used to extract features from the resulting partitions. For example, identify a specific percentile in all the rows, then calculate different statistical indicators from the percentiles. Alternatively, calculate a statistical indicator from all the values in the partitions. After this step, the original $m$ signals of an instance are represented in an $m \times c \times n \times g$ dimensional feature space, where $n$ is the number of extraction methods used, and $g$ is the number of $(r, l, k)$ triplets used.

\item[\textbf{[B1]}] \textbf{Classifier Tuning and Evaluation.} Utilize new features to tune advanced classifiers (e.g., $K$-nearest neighbors) via Bayesian hyperparameter optimization and cross-validation. Evaluate classifiers' accuracy, tuning, and classification time on the training set.

\item[\textbf{[B2]}] \textbf{Test and Benchmark.} Similar to steps [A5--A6], transform the test set ($Te$), generate new features, and apply tuned classifiers. Measure out-of-sample classification speed and accuracy. Benchmark results against state-of-the-art methods.

\end{enumerate}
\end{adjustwidth}

\begin{figure}[hbt!]
\caption{Applied ML framework}
\label{F:1}
 \centering
  \includegraphics[width=0.4\linewidth]{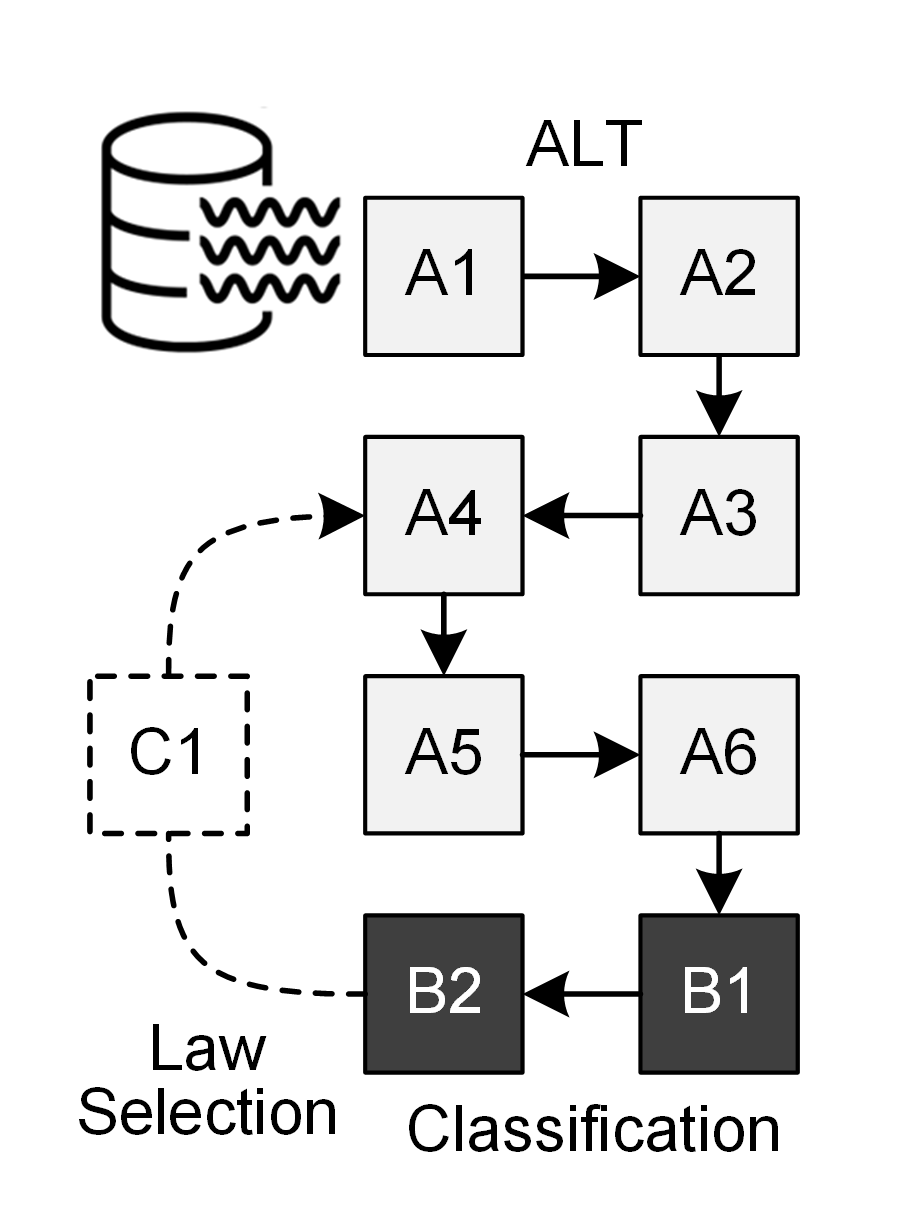}
\end{figure}

Figure \ref{F:1} illustrates the complete feature representation and classification procedure, including a law selection step [C1] that is planned for implementation in a future study---see Section \ref{sec:conclusions} for more details.

\noindent
\subsection{Software and Settings}
\label{sec:2.4}

\noindent
We implemented steps [A1–A6] in Python to transform the original feature spaces. The transformed features were then used to train KNN~\cite{KNN} and SVM~\cite{SVM} classifiers in the MATLAB Classification Learner App to perform steps [B1–B2].\footnote{More information can be found at \url{https://www.mathworks.com/help/stats/classificationlearner-app.html} (retrieved: January 15, 2025).} During the classification procedure, a 30-step Bayesian hyperparameter optimization with 5-fold cross-validation was applied.\footnote{As a benchmark, we also used optimizable neural networks on the raw time-series data with a 500-step Bayesian hyperparameter optimization and 5-fold cross-validation. These benchmark results are presented in Table~\ref{tabl:benchmark} in the Appendix.}

We also optimized the hyperparameters of the proposed method \((r, l, k)\) to achieve the highest classification accuracy. Furthermore, during feature extraction~[A6], we incorporated various statistical indicator pairs that yielded the best performance. From the rows of the matrix \(O\), the mean and 5\textsuperscript{th} percentile were computed, followed by calculations of the mean, variance, and the third and fourth moments. The exact parameter settings applied to each dataset are detailed in Table~\ref{tabl:parameters} in the Appendix.

\section{Results and discussion}
\label{sec:results}

\noindent
The classification outcomes obtained with ALT are summarized in Table~\ref{tabl:results}.

\begin{table*}[ht]
    \centering
    \caption{Classification results}
    \label{tabl:results}
    \resizebox{0.75\textwidth}{!}{%
    \begin{tabular}{lcccrrc}
        \toprule
       \multicolumn{1}{c}{\textbf{Dataset}} & 
        \multicolumn{1}{c}{\shortstack{\textbf{Validation} \\ \textbf{Accuracy}}} & 
        \multicolumn{1}{c}{\shortstack{\textbf{Test} \\ \textbf{Accuracy}}} & 
        \multicolumn{1}{c}{\shortstack{\textbf{Classification} \\ \textbf{Method}}} & 
        \multicolumn{1}{c}{\shortstack{\textbf{Transform.} \\ \textbf{Time (s)}}} & 
        \multicolumn{1}{c}{\shortstack{\textbf{Classification} \\ \textbf{Time (s)}}} & 
        \multicolumn{1}{c}{\shortstack{\textbf{Benchmark}}} \\
        \midrule
        BasicMotions             & 100.0\%             & 100.0\%       & KNN                   & 1.90               & 9.00                  & 95.3--100.0\% \\ 
        Coffee                   & 100.0\%             & 100.0\%       & KNN                   & 1.22               & 6.41                  & 78.6--100.0\% \\
        Epilepsy                 & 96.1\%             & 97.8\%       & SVM                   & 84.58              & 12.87                 & 85.0--100.0\% \\
        Epilepsy2                & 95.0\%             & 93.8\%       & KNN                   & 48.09              & 7.43                  & 89.4--100.0\% \\
        FordA                    & 97.5\%             & 97.5\%       & SVM                   & 915.80             & 28.95                 & 49.0--100.0\% \\
        FordB                    & 84.9\%             & 94.4\%       & KNN                   & 3069.00            & 14.54                 & 50.9--100.0\% \\
        GunPoint1                & 100.0\%             & 96.7\%       & SVM                   & 7.49               & 7.90                  & 68.0--100.0\% \\
        GunPoint2                & 98.5\%             & 93.0\%       & SVM                   & 16.25              & 13.96                 & 57.0--100.0\% \\
        GunPoint3                & 100.0\%             & 99.4\%       & KNN                   & 6.99               & 6.85                  & 68.0--100.0\% \\
        GunPoint4                & 100.0\%             & 100.0\%       & KNN                   & 2.32               & 6.62                  & 88.0--100.0\% \\
        PowerCons                & 92.4\%             & 93.3\%       & SVM                   & 3.45               & 9.07                  & 73.0--100.0\% \\
        \bottomrule
        \multicolumn{7}{p{0.95\textwidth}}{\justifying \scriptsize
        \noindent
        \textbf{Note:} The original names of the GunPoint datasets, marked by numbers, are as follows: 1. GunPoint; 2. GunPointAgeSpan; 3. GunPointMaleVersusFemale; 4. GunPointOldVersusYoung. Results were obtained using 30 iterations of Bayesian hyperparameter optimization in the MATLAB Classification Learner App. Benchmarks were derived from the test accuracies reported by the studies summarized in Table \ref{tabl:existing_benchmark} in the Appendix.}
    \end{tabular}%
    }
\end{table*}

As Table \ref{tabl:results} shows, ALT consistently achieves high validation and test accuracies across all eleven datasets, including perfect scores (100\%) on \textit{BasicMotions}, \textit{Coffee}, and \textit{GunPoint4}. Transformations typically complete within a practical time frame; however, for larger datasets (e.g., \textit{FordB}), the transformation step can be more time-consuming. This overhead arises primarily from shapelet vector generation and spectral decomposition steps. Once the transformed features are computed, classification (via KNN or SVM) is relatively fast.

Table~\ref{tabl:parameters} details the hyperparameter and feature-extraction settings employed for each experiment, including the ratio of data used for shapelet generation versus classifier training. Notably, only a small subset of the data is typically required for learning shapelets, highlighting ALT’s efficiency in deriving class-relevant patterns.

For additional context, Table~\ref{tabl:benchmark} compares ALT’s accuracy to that of a neural network benchmark using an optimizable feed-forward architecture (MLP) implemented in MATLAB on the raw time-series data.\footnote{Neural networks were tuned using $500$-step Bayesian hyperparameter optimization and $5$-fold cross-validation.} On most datasets, regardless of their length, ALT outperforms or closely matches the neural network solution despite having far fewer hyperparameters and a shorter optimization process. Furthermore, the benchmark compilation in Table~\ref{tabl:results} demonstrates that ALT is highly competitive against a wide range of state-of-the-art approaches, including shapelet-based methods and advanced neural and kernel techniques.

Across tasks, ALT’s ability to capture subsequence patterns of varying lengths proves advantageous, particularly for datasets with subtle class-distinguishing events (e.g., \textit{Epilepsy}, \textit{GunPoint2}). This adaptability is reflected in consistent improvements over baseline neural methods, which often struggle with more complex sensor signals (e.g., \textit{FordA}, \textit{FordB}). Although certain tasks (e.g., \textit{Coffee}, \textit{GunPoint4}) are relatively straightforward for most algorithms, ALT maintains robust reliability while retaining interpretability by design.

\section{Conclusion and Future Works}
\label{sec:conclusions}

\noindent
In this paper, we introduced ALT, a novel method for time series classification that generalizes our previous LLT approach. By incorporating variable-length shifted windows, it captures local subsequence patterns of different scales and embeds them in a linearly separable feature space. Extensive experiments across eleven diverse datasets confirm ALT’s capacity to deliver competitive or state-of-the-art results, as evidenced by Tables~\ref{tabl:results}, \ref{tabl:benchmark}, and~\ref{tabl:existing_benchmark}.

In future work, we plan to integrate data-driven mechanisms for automatically tuning $(r,l,k)$, thus further reducing manual hyperparameter exploration. Additionally, we aim to investigate shapelet pruning techniques (see step [C] in Figure \ref{F:1}) to lower computational overhead, making ALT scalable to very large time series with minimal performance loss. The method’s interpretability could also be enriched by qualitative visualization of extracted shapelet vectors, potentially illuminating latent domain structures. Finally, exploring ALT’s capabilities in specialized domains like multi-channel EEG monitoring or IoT anomaly detection may reveal further performance gains and highlight the role of domain-specific knowledge in shaping the transformation pipeline.

\section*{Acknowledgments}

\noindent
The research was supported by the Hungarian Government and the European Union in the framework of a Grant Agreement No. MILAB RRF-2.3.1-21-2022-00004. Project no. PD142593 was implemented with the support provided by the Ministry of Culture and Innovation of Hungary from the National Research, Development, and Innovation Fund, financed under the PD\_22 ``OTKA'' funding scheme.

\bibliographystyle{model5-names}
\bibliography{refs}

\appendix
\renewcommand{\thetable}{A.\arabic{table}}
\setcounter{table}{0}
\renewcommand\theHtable{Appendix.\thetable}
\section*{Appendix}

\begin{table}[ht]
    \centering
    \caption{Classification of raw datasets with neural networks}
    \resizebox{0.45\textwidth}{!}{
    \label{tabl:benchmark}
    \begin{tabular}{cccc}
        \toprule
        \textbf{Dataset} & \textbf{Validation accuracy} & \textbf{Test accuracy} & \textbf{Training time (s)} \\
        \midrule
        BasicMotions & 72.5\% & 87.5\% & 1941.7 \\
        Coffee & 100.0\% & 100.0\% & 1105.7 \\
        Epilepsy & 65.7\% & 67.4\% & 2745.3 \\
        Epilepsy2 & 80.0\% & 89.9\% & 1248.9 \\
        FordA & 72.7\% & 72.0\% & 7347.9 \\
        FordB & 63.0\% & 66.0\% & 5211.7 \\
        GunPoint1 & 98.0\% & 94.0\% & 1380.7 \\
        GunPoint2 & 96.3\% & 98.1\% & 2023.2 \\
        GunPoint3 & 99.3\% & 99.7\% & 1513.2 \\
        GunPoint4 & 100.0\% & 100.0\% & 866.5 \\
        PowerCons & 100.0\% & 98.9\% & 1300.9 \\
        \bottomrule
        \multicolumn{4}{p{0.6\textwidth}}{\justifying \scriptsize
        \noindent
        \textbf{Note:} The original names of the GunPoint datasets, marked by numbers, are as follows: 1. GunPoint; 2. GunPointAgeSpan; 3. GunPointMaleVersusFemale; 4. GunPointOldVersusYoung. Results were obtained using 500 iterations of Bayesian hyperparameter optimization and 5-fold cross-validation in the MATLAB Classification Learner App.}
    \end{tabular}
    }
\end{table}

\begin{table*}[ht]
    \centering
    \caption{Applied parameters}
    \resizebox{0.9\textwidth}{!}{
    \label{tabl:parameters}
    \begin{tabular}{cccc}
        \toprule
        \textbf{Dataset} & \textbf{\vtop{\hbox{\strut Learn-}\hbox{\strut train ratio}}} & \textbf{Method} & \textbf{Used $(r, l, k)$ values} \\
        \midrule
        BasicMotions & 0.25 & mean - mean, \(5^{\text{th}}\) percentile - \(4^{\text{th}}\) moment & (53, 27, 1) \\
        Coffee & 0.25 & \(5^{\text{th}}\) percentile - mean & (3, 2, 1) \\
        Epilepsy & 0.25 & mean - mean & (29, 15, 1), (69, 35, 1), (89, 45, 1), (149, 75, 1), (169, 85, 1), (189, 95, 1) \\
        Epilepsy2 & 0.25 & \(5^{\text{th}}\) percentile - mean, \(5^{\text{th}}\) percentile - variance & (19, 10, 1), (29, 15, 1) \\
        FordA & 0.20 & \(5^{\text{th}}\) percentile - mean & (23, 12, 1), (29, 15, 1), (85, 43, 1), (95, 48, 1), (205, 103, 1)\\
        FordB & 0.50 & \(5^{\text{th}}\) percentile - mean & (19,10,1),(39,20,1),(129,65,1),(139,70,1),(159,80,1), (169,85,1), \\
        &&& (179,90,1),(199,100,1),(209,105,1),(275,138,1) \\
        GunPoint1 & 0.20 & mean - mean & (7, 4, 1), (31, 2, 1), (51, 6, 1), (81, 6, 1), (121, 11, 1), (121, 31, 1), \\
        &&& (121, 61, 1), (121, 5, 1) \\
        GunPoint2 & 0.50 & mean - mean, \(5^{\text{th}}\) percentile - excess kurtosis & (49, 25, 1), (59, 30, 1), (69, 35, 1), (89, 45, 1) \\
        GunPoint3 & 0.20 & mean - mean, \(5^{\text{th}}\) percentile - mean & (3, 2, 1), (19, 10, 1), (39, 20, 1), (109, 55, 1) \\
        GunPoint4 & 0.50 & mean - mean & (3, 2, 1) \\
        PowerCons & 0.20 & mean - mean & (3, 2, 1), (99, 50, 1) \\
        \bottomrule
        \multicolumn{4}{p{1.2\textwidth}}{\justifying \scriptsize
        \noindent
        \textbf{Note:} The original names of the GunPoint datasets, marked by numbers, are as follows: 1. GunPoint; 2. GunPointAgeSpan; 3. GunPointMaleVersusFemale; 4. GunPointOldVersusYoung. Results were obtained using 30 iterations of Bayesian hyperparameter optimization in the MATLAB Classification Learner App.}
    \end{tabular}
    }
\end{table*}

\begin{table*}[ht]
    \centering
    \caption{Literature benchmarks}
    \resizebox{0.8\textwidth}{!}{
    \label{tabl:existing_benchmark}
    \begin{tabular}{cccc}
        \toprule
        \textbf{Database} & \textbf{Test accuracy (\%)} & \textbf{Reference} & \textbf{Method} \\
        \midrule
        BasicMotions & 95.3--100.0 & \cite{artic_letime_series_classification_bake_off_Pasos_Alejandro} & DTWD, ROCKET, CIF, HIVE-COTE \\
        Coffee & 96.0--100.0 & \cite{dhariyal2023back} & RandomForest, Rocket, Minirocket, Multirocket \\
        Coffee & 78.6--100.0 & \cite{10.6703--2077100067} & Raw-ResNet, FoldCount-1NN, TimeAxisArea-1NN, DWT-1NN \\
        Epilepsy & 96.3--100.0 & \cite{artic_letime_series_classification_bake_off_Pasos_Alejandro} & DTWD, ROCKET, CIF, HIVE-COTE \\
        Epilepsy & 95.7--97.1 & \cite{10446381} & Debiased Contrastive Learning with Weak Supervision \\
        Epilepsy & 85.0--99.0 & \cite{Hussein_Dina_Nelson_Lubah_Bhat_Ganapati_Sensor_Aware_Classifiers} & CNN \\
        Epilepsy2 & 89.4--100.0 & \cite{lin2023nutime} & Multi-Scaled Embedding for Large-Scale Time-Series Pretraining \\
        FordA & 96.8--100.0 & \cite{mukhopadhyay2024time} & Lightweight Attention Networks \\
        FordA & 79.3--86.4 & \cite{xi2023lb} & LB-SimTSC (Similarity-Aware Graph Neural Network) \\
        FordA & 49.0--95.0 & \cite{dhariyal2023back} & RandomForest, Rocket, Minirocket, Multirocket \\
        FordA & 74.54--95.6 & \cite{doi:10.1137/1.9781611978032.91} & LSRSC (Centered Kernel Alignment) \\
        FordA & 56.7--93.6 & \cite{10.6703--2077100067} & Raw-ResNet, FoldCount-1NN, TimeAxisArea-1NN, DWT-1NN \\
        FordA & 53.4--71.3 & \cite{ceni2023residual} & Residual Reservoir Computing Neural Networks \\
        FordA & 89.0 & \cite{10.1007/978-3-031-44070-0_9} & Convolutional Neural Networks \\
        FordA & 96.5 & \cite{bostrom2018shapelet} & Shapelet Transform \\
        FordA & 50.6--90.9 & \cite{eldele2023self} & Time-Series/Class-Aware Temporal and Contextual Contrasting \\
        FordB & 92.9--100.0 & \cite{mukhopadhyay2024time} & Lightweight Attention Networks \\
        FordB & 49.0--83.0 & \cite{dhariyal2023back} & RandomForest, Rocket, Minirocket, Multirocket \\
        FordB & 63.8--83.1 & \cite{doi:10.1137/1.9781611978032.91} & LSRSC (Centered Kernel Alignment) \\
        FordB & 53.1--81.7 & \cite{10.6703--2077100067} & Raw-ResNet, FoldCount-1NN, TimeAxisArea-1NN, DWT-1NN \\
        FordB & 51.9--56.4 & \cite{ceni2023residual} & Residual Reservoir Computing Neural Networks \\
        FordB & 70.0 & \cite{10.1007/978-3-031-44070-0_9} & Convolutional Neural Networks \\
        FordB & 91.5 & \cite{bostrom2018shapelet} & Shapelet Transform \\
        FordB & 50.9--88.2 & \cite{eldele2023self} & Time-Series/Class-Aware Temporal and Contextual Contrasting \\
        GunPoint1 & 85.0--100.0 & \cite{dhariyal2023back} & RandomForest, Rocket, Minirocket, Multirocket \\
        GunPoint1 & 85.0--100.0 & \cite{dhariyal2023back} & RandomForest, Rocket, Minirocket, Multirocket \\
        GunPoint1 & 68.0--99.0 & \cite{10.6703--2077100067} & Raw-ResNet, FoldCount-1NN, TimeAxisArea-1NN, DWT-1NN \\
        GunPoint2 & 57.0--100.0 & \cite{dhariyal2023back} & RandomForest, Rocket, Minirocket, Multirocket \\
        GunPoint3 & 68.0--100.0 & \cite{dhariyal2023back} & RandomForest, Rocket, Minirocket, Multirocket \\
        GunPoint4 & 88.0--100.0 & \cite{dhariyal2023back} & RandomForest, Rocket, Minirocket, Multirocket \\
        PowerCons & 73.0--100.0 & \cite{dhariyal2023back} & RandomForest, Rocket, Minirocket, Multirocket \\
        \bottomrule
        \multicolumn{4}{p{1.1\textwidth}}{\justifying \scriptsize
        \noindent
        \textbf{Note:} The original names of the GunPoint datasets, marked by numbers, are as follows: 1. GunPoint; 2. GunPointAgeSpan; 3. GunPointMaleVersusFemale; 4. GunPointOldVersusYoung.}
    \end{tabular}
    }
\end{table*}

\end{document}